\documentclass[conference]{IEEEtran}
\IEEEoverridecommandlockouts
\usepackage{cite}
\usepackage{amsmath,amssymb,amsfonts}
\usepackage{algorithmic}
\usepackage{graphicx}
\usepackage{textcomp}
\usepackage{xcolor}
\usepackage{xurl}
\usepackage{float}
\usepackage{stfloats} 
\usepackage{mathptmx}

\usepackage{fancyhdr}

\fancypagestyle{firstpage}{
  \fancyfoot[L]{\fontsize{8}{10}\selectfont 979-8-3503-9478-8/24/\$31.00 ©2024 IEEE}
}

\begin{document}

\title{Parameter Efficient Diverse Paraphrase Generation Using Sequence-Level Knowledge Distillation}

\author{\IEEEauthorblockN{Lasal Jayawardena}
\IEEEauthorblockA{\textit{School of Computing} \\
\textit{Informatics Institute of Technology}\\
Colombo, Sri Lanka \\
lasal.20200344@iit.ac.lk}
\and
\IEEEauthorblockN{Prasan Yapa}
\IEEEauthorblockA{\textit{School of Computing} \\
\textit{Informatics Institute of Technology}\\
Colombo, Sri Lanka \\
prasan.y@iit.ac.lk}

}

% \author
% {
%   \IEEEauthorblockN{Anonymous}
%   \IEEEauthorblockA{\textit{Removed for Anonymity}
%   \\ \textit{Removed for Anonymity} 
%   \\ Removed for Anonymity}
%   \and
%   \IEEEauthorblockN{Anonymous}
%   \IEEEauthorblockA{\textit{Removed for Anonymity} 
%   \\ \textit{Removed for Anonymity} 
%   \\ Removed for Anonymity}
% }

\maketitle

\begin{abstract}
Over the past year, the field of Natural Language Generation (NLG) has experienced an exponential surge, largely due to the introduction of Large Language Models (LLMs). These models have exhibited the most effective performance in a range of domains within the Natural Language Processing and Generation domains. However, their application in domain-specific tasks, such as paraphrasing, presents significant challenges. The extensive number of parameters makes them difficult to operate on commercial hardware, and they require substantial time for inference, leading to high costs in a production setting.
In this study, we tackle these obstacles by employing LLMs to develop three distinct models for the paraphrasing field, applying a method referred to as sequence-level knowledge distillation. These distilled models are capable of maintaining the quality of paraphrases generated by the LLM. They demonstrate faster inference times and the ability to generate diverse paraphrases of comparable quality. A notable characteristic of these models is their ability to exhibit syntactic diversity while also preserving lexical diversity, features previously uncommon due to existing data quality issues in datasets and not typically observed in neural-based approaches.
Human evaluation of our models shows that there is only a 4\% drop in performance compared to the LLM teacher model used in the distillation process, despite being 1000 times smaller. This research provides a significant contribution to the NLG field, offering a more efficient and cost-effective solution for paraphrasing tasks. 
\end{abstract}


\section{Introduction}
Paraphrase Generation, alternatively known as Question Paraphrase Generation, occupies a key role as a fundamental task within the field of Natural Language Processing (NLP). For several decades, this section of the field has been explored consistently and the derived results have found immense applications in enhancing data augmentation processes\cite{mckeown_paraphrasing_1979}. The influence of Paraphrase Generation on data augmentation is pivotal for optimizing the output of numerous NLP operations, which leads to a substantial enrichment of training data \cite{meteer_strategies_1988}. Throughout the years, extensive research has gone into this division of NLP, generating key strategies for optimizing the process \cite{kozlowski_generation_2003}. 

Traditional methods of paraphrase generation, such as rule-based \cite{lin_discovery_2001}, thesaurus-based \cite{kauchak_paraphrasing_2006}, and SMT-based methods \cite{wubben_paraphrase_2010}, have been widely used in the past. However, they are inherently limited in their capacity to generate diverse and contextually accurate paraphrases, therefore recent research has been mainly based on neural-based approaches.

The advent of Large Language Models (LLMs) has significantly transformed the landscape of NLP, including the domain of paraphrase generation. These models, underpinned by advanced neural networks such as Transformers, have shown unparalleled performance in a spectrum of NLP applications, covering everything from categorizing texts and assessing sentiments to translating languages and textual content generation. \cite{zhao_survey_2023}. LLMs undergo training using extensive text corpora and web resources, which equips them with the capability to decipher complex structures and patterns inherent in human language. Advanced techniques such as Reinforcement Learning with Human Feedback (RLHF) \cite{li_reinforcement_2023} have drastically improved its capabilities. This extensive training allows them to generate human-like writing, which is not just syntactically accurate but also pertinent and logically consistent within the context. Moreover, LLMs can understand and generate text in a context-dependent manner, an essential element in the creation of paraphrases.

Despite the remarkable proficiency of LLMs, several challenges remain. One of the most significant issues is the high number of parameters in these models. The sheer size of LLMs makes them difficult to run on consumer hardware, limiting their accessibility for many users and applications. Additionally, the large number of parameters also leads to longer inference times, which can be a bottleneck in real-time applications. These challenges highlight the urgent need for domain-specific models that maintain the high performance of LLMs while being more efficient and accessible. In the case of paraphrase generation, there is a need for models that can generate high-quality paraphrases quickly and efficiently, without the need for high-end hardware.

\thispagestyle{firstpage}

In response to this need, we have leveraged the power of LLMs to distill and build models that are significantly smaller than the original teacher models. Our models are approximately a thousand times smaller in terms of the number of parameters, making them much more efficient and easier to run on consumer hardware. Despite their smaller size, these models maintain the high-quality paraphrase generation capabilities of their larger counterparts, being a testament to the effectiveness of our approach. As shown in Section \ref{sec:evaluation-section}, we have utilized a gold standard evaluation strategy which is not commonly seen in most paraphrase generation research \cite{zhou_paraphrase_2021}. In addition, results seen in Section \ref{sec:huma-eval-sec}, qualitative results obtained by employing human evaluators, illustrate that the distilled models were indeed capable of maintaining the quality and diversity of the output despite being drastically smaller than the LLM it is trained on. 

The results of our research not only contribute to the field of paraphrase generation but also demonstrate the potential of knowledge distillation as a strategy for leveraging the power of LLMs in a more efficient and accessible manner. We believe that our work lays the groundwork for upcoming studies within the field of paraphrase generation, opening up new possibilities for the application of LLMs effectively. 

\section{Related Work}

\subsection{Paraphrase Datasets}
To support the exploration and creation of models for generating paraphrases, a variety of datasets have been established. The Paraphrase Database (PPDB) \cite{ganitkevitch_ppdb_2013} is a comprehensive resource that contains over 220 million paraphrase pairs. However, its utility has been questioned due to its exclusive focus on phrasal and lexical paraphrases, neglecting sentence paraphrases.

The Twitter URL dataset \cite{lan_continuously_2017} is a large-scale collection of sentential paraphrases sourced from Twitter. However, due to the noisiness of the labels, this dataset is not widely used. The Wiki Answer dataset \cite{fader_paraphrase-driven_2013} contains an estimated 18 million pairs of paraphrased questions, but it is limited in scope as all the sentences provided are in the form of questions.

The MSCOCO dataset \cite{fleet_microsoft_2014} was primarily characterized as a comprehensive object detection dataset. It comprises over 120,000 images, each of which is accompanied by five distinct captions, contributed by five separate annotators. The Microsoft Research Paraphrase Corpus (MRPC) Dataset \cite{dolan_automatically_2005} comprises 5800 sentence pairs derived from online news sources. It also includes human annotations that denote whether each pair represents a paraphrase or semantic equivalent.

The Quora Dataset or Quora Question Pair Dataset \cite{iyer_first_2017} contains 150,000 question pairs that are annotated as paraphrases. The ParaNMT dataset \cite{wieting_paranmt-50m_2018} comprises over 50 million pairs of English sentential paraphrases. These pairs were independently created by utilizing bidirectional translation for converting the non-English elements within a significant Czech-to-English equivalent dataset.

In the development of the ParaBank datasets, a Czech-English Neural Machine Translation (NMT) system was employed to create new variations of English reference sentences \cite{hu_parabank_2019,hu_large-scale_2019}. The PAWS Dataset \cite{zhang_paws_2019} contains sentences with high bag-of-words (BOW) overlap but having different word order.

Despite the variety of datasets available, there are still challenges in using them for paraphrase generation. The noise introduced in these datasets due to utilizing techniques such as back-translation can lead to error propagation, which cannot be mitigated even by improving the architecture of the model. Therefore, a data-centric approach is needed to handle these issues.

\subsection{Paraphrase Generation}
Paraphrase generation has seen significant advancements with the introduction of sophisticated techniques. These include multi-round generation \cite{lin_pushing_2021}, which involves generating multiple iterations of paraphrases, and reinforcement learning-based paraphrasing \cite{liu_learning-exploring_2020}, which employs reinforcement learning algorithms to optimize the paraphrasing process. Another noteworthy approach is prompt-tuning \cite{chowdhury_novelty_2022}, which fine-tunes the prompts used in the generation process to yield better paraphrases. There is also a subset of research dedicated to enhancing the syntactic diversity of the generated paraphrases. This is achieved through various methods such as sampling from latent spaces \cite{cao_divgan_2020}, which generates diverse paraphrases by sampling different points in the latent space, and controlling the word order \cite{goyal_neural_2020}, which manipulates the arrangement of words to create diverse paraphrases. However, these methods often focus on one aspect of diversity, typically neglecting lexical diversity. Furthermore, the quality of data used in these approaches often leaves room for improvement.

Generative adversarial networks (GANs) \cite{goodfellow_generative_2014} have also been utilized in paraphrase creation. The distinct characteristics of textual content generation pose an obstacle to the conventional training techniques applied in GANs. To overcome this, the concept of policy gradient is employed \cite{yu_seqgan_2017}. Word-level paraphrasing is another technique that focuses on generating paraphrases by substituting original words with synonyms. Some researchers have leveraged external linguistic knowledge to achieve this \cite{lin_integrating_2020}, while others have proposed unique mechanisms to learn synonym mappings \cite{fu_paraphrase_2020}. These mechanisms can significantly enhance lexical diversity.

In the search for more sophisticated paraphrase generation, experts have investigated techniques to manage the syntactic structure of paraphrased text, integrating various levels of detail. Studies in this field typically fall into two groups based on their approach to syntax control: implicit and explicit. In explicit control strategies, the sentence's syntactic tree is transformed into vector forms, which are then incorporated into the decoder at each step of decoding \cite{kumar_syntax-guided_2020}. Implicit control methods, alternatively, acquire knowledge of syntactic information distribution through a Variational Autoencoder (VAE). In this approach, a syntax variable, drawn from the learned distribution, is integrated into the decoder at every step of decoding. \cite{chen_semantically_2020}. 

Building on these syntax-focused methods, multi-level paraphrasing techniques combine multiple granularity levels, enabling their models to generate synonyms, substitute phrases, and rearrange sentential structures \cite{kazemnejad_paraphrase_2020}. These techniques aim to create a more comprehensive and nuanced approach to paraphrase generation.

However, a common challenge faced by most of these approaches stems from the scarcity of extensive corpora containing significant quality paraphrases. A high-quality paraphrase should be lexically diverse, syntactically diverse, grammatically correct, and semantically similar. Balancing all these aspects in paraphrase generation remains a significant challenge.

\subsection{Knowledge Distillation}

The technique of knowledge distillation is utilized to educate compact models, frequently termed as the student, through leveraging the ``knowledge" from a more expansive model, designated as the teacher \cite{hinton_distilling_2015}. A common method of knowledge distillation is to train the student with an additional goal of aligning with the teacher's representation, such as logits, output probability, or intermediate activation \cite{jiao_tinybert_2020}.

In the context of sequence-to-sequence or generative models, \cite{kim_sequence-level_2016} introduced the idea of sequence-level knowledge distillation. This method involves the generation of a synthetic output by conducting inference with the teacher model, which is subsequently used to train the learner/student model. The efficiency of sequence-level distillation lies in the fact that it only necessitates running the typically large teacher model once. The efficacy of sequence-level distillation has been demonstrated in previous studies \cite{bogoychev_edinburghs_2020}. Recent research has adopted this technique, particularly with the use of LLMs \cite{wu_lamini-lm_2023}. Further extensions of this technique have been explored, such as the use of reverse Kullback-Leibler divergence (KLD) objectives to enhance the distillation process \cite{gu_knowledge_2023}. This indicates a growing interest in the field of NLP in utilizing sequence-level distillation to develop smaller, yet effective models, especially with the recent research involving LLMs.

\section{Methodology}

\begin{figure*}
\centering
% \hspace{-1.2cm} 
% \includegraphics[width=0.99\textwidth]{High Level Diagram.jpeg} 
% \includegraphics[width=0.99\textwidth]{Model Zoo High Level Diagram .jpeg}
% \includegraphics[width=1.055\textwidth]{high_level_diagram.png}
\includegraphics[width=1.0\textwidth]{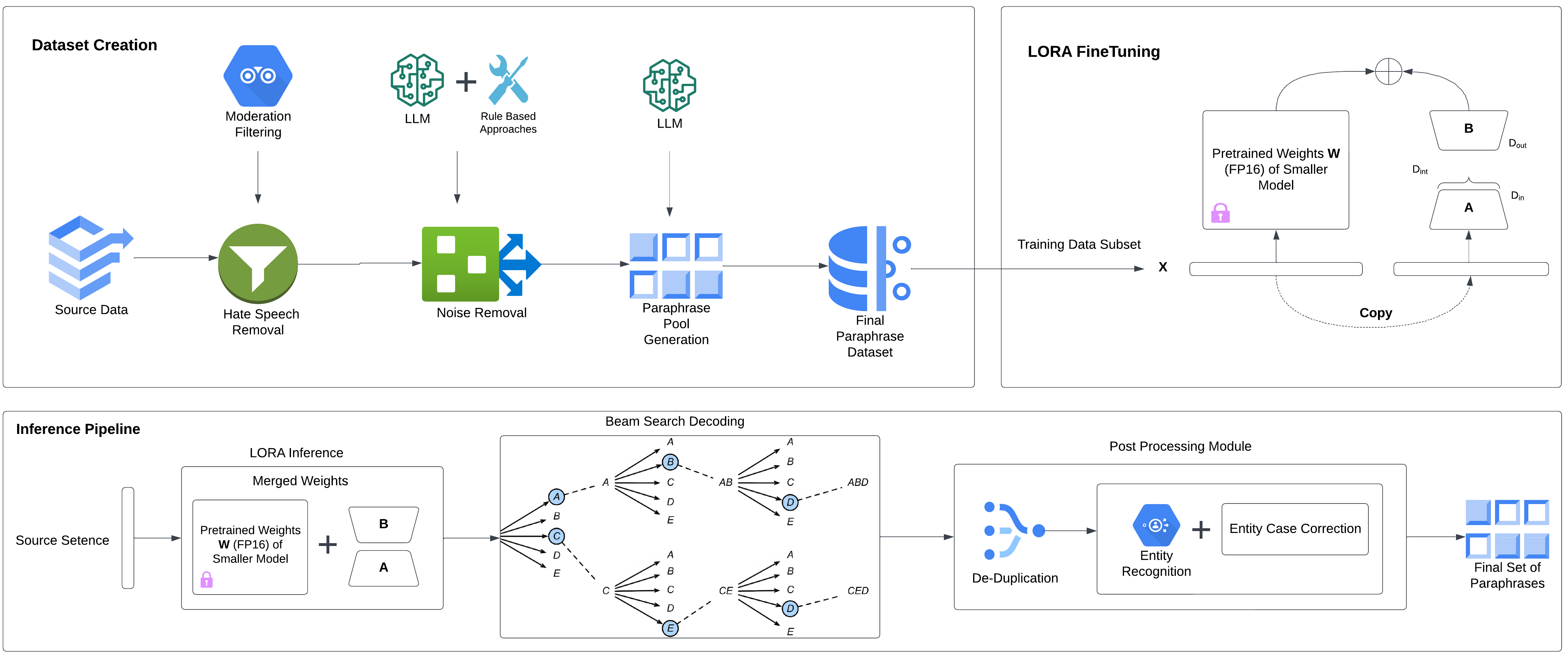}
% \caption{Model Training Architecture}
\caption{High-Level Architecture Diagram for Training and Inference Phases.}
\label{fig}
\end{figure*}

We utilize a data-centric sequence-level knowledge distillation technique where we utilize the ChatGPT (gpt-3.5-turbo) LLM to create smaller models which are capable of producing diverse paraphrases to a quality similar but with a significantly less number of parameters. The detailed methodology will be broken down and explained in the subsequent sections.

\subsection{Dataset Creation}
% \subsubsection{Data Sources}
To create the full dataset we used multiple data sources. We incorporated a subset of the Quora Dataset \cite{dolan_automatically_2005}, specifically choosing sentence pairs labeled as paraphrases. Next, we utilized PAWSWiki, a segment within the PAWS (Paraphrase Adversaries from Word Scrambling) Dataset\cite{zhang_paws_2019}. However, we deliberately avoided using PAWSQQP as it contained source sentences identical to those in the Quora Dataset. These two are the main corpora employed in the training process. For evaluation purposes, the Microsoft Research Paraphrase Corpus (MRPC) Dataset \cite{dolan_automatically_2005}, the MSCOCO dataset \cite{fleet_microsoft_2014}, the Wiki Answer dataset \cite{fader_paraphrase-driven_2013} and the Twitter URL dataset \cite{lan_continuously_2017} was incorporated.
% \subsubsection{Data Generation}
% \subsubsection{Additonal Processing}

For the paraphrase pair generation phase, we initially combined the aforementioned data sources, selecting approximately 750,000 source sentences for paraphrase generation. An initial pass was conducted to filter out offensive content, using OpenAI’s Moderation Endpoint. This process flagged any source sentence that fell under the categories of offensive content, allowing us to filter out approximately source sentences containing such content. In addition to this, data sources created using back-translation often contain a significant amount of noise, including non-English source sentences. To tackle this, we developed a new prompt for the gpt-3.5-turbo model, instructing it to identify English sentences and generate paraphrases, or output ”Error” for non-English sentences. This method, coupled with a rule-based approach to filter out certain responses from the model, proved efficient in significantly reducing noise.

We then used the ChatGPT (gpt-3.5-turbo) LLM to augment the source sentences. The prompt used for each dataset varied slightly. For paraphrase generation, we adjusted the temperature parameter value to zero. The model output was a string containing augmented, diverse sentence paraphrases in a numbered list format. This string was subsequently processed to generate a list of strings, which was then used to construct a pool of paraphrases. By utilizing the paraphrase pool, we successfully generated nearly 2 million unique sentence paraphrase pairs. This was then used for model training and evaluation.

\subsection{Model Training}
% \subsubsection{Model Architectures}

The models selected for distillation were T5-small \cite{raffel_exploring_2019}, Flant5-small\cite{chung_scaling_2022}, and BART-base\cite{lewis_bart_2019}, each chosen for their unique strengths and capabilities. These models are built on an encoder-decoder architecture, a design that is particularly effective for tasks involving conditional generation, such as paraphrase generation.

The T5-small model is known for its efficiency and high performance in text generation tasks. The model is crafted to adeptly deal with a wide range of NLP tasks, making it a versatile choice for our project. Its small size, compared to larger versions of T5, makes it more computationally efficient while still delivering strong performance. The T5-small model has only around 60 million parameters.

The Flant5-small model is a multilingual version of T5. It has been trained on multiple languages, which could potentially enhance the quality of paraphrase generation by leveraging the syntactic and semantic similarities across languages. By including this model in our study, we aimed to explore the potential benefits of multilingual capabilities in paraphrase generation. The Flant5-small is slightly bigger than the T5-small model with around 80 million parameters.

Lastly, the BART-base model was included because of its impressive capabilities in generating text and summarizing information. BART stands out in its pretraining methodology, being designed to restore the original content after specific tokens are obscured or complete sentences are rearranged. This makes BART particularly effective at tasks that require understanding the broader context of a sentence, such as paraphrase generation. This model is slightly bigger with around 140 million parameters which is still a thousand times smaller than ChatGPT.

By distilling these three models, we aimed to leverage their unique strengths and capabilities to create a robust and efficient model for paraphrase generation.

% \subsubsection{Training Setup}
% \subsubsection{Hyperparameters}
For the training process, we utilized the Quora Question Pairs and PAWS datasets. To make the training more effective we lower cased the data and trained the models to further simplify the learning process so that models could learn on the core task effectively. The training method employed was Low-Rank Adaptation (LoRA) \cite{hu_lora_2021}. This method preserves the weights of the pre-trained model and integrates trainable rank decomposition matrices into every layer of the Transformer structure. Consequently, this strategy considerably lowers the number of parameters needed for subsequent tasks, resulting in reduced GPU memory usage and enhanced training throughput. The LoRA technique was applied through the use of a library specifically designed for Parameter-Efficient Fine-Tuning (PEFT), developed by Hugging Face\footnote{https://github.com/huggingface/peft}. The training of each model was conducted using RTX A100 40GB GPU, using approximately 1.4 million rows of data. This approach is very effective in knowledge distillation when working with such a large volume of data.

The hyperparameters for the three models were set as follows:
The models were programmed to accommodate a diverse range of sentence lengths by setting the maximum sequence length at 256. The models were trained for 10 epochs, providing a balance between training time and model performance.  To ensure numerical stability, the Adam optimizer was utilized, accompanied by an epsilon value fixed at 1e-08. A learning rate of 0.0003 was determined, along with a maximum gradient norm limited to 1.0, which serves to inhibit excessively large gradients. The LoRA configuration for these models was set with a rank (r) of 8 and an alpha value of 32, providing a balance between model complexity and performance. To avert the risk of overfitting, the dropout rate was established at 0.1. In all models, ``paraphrase" was used as the prefix for the input sentences during training. The training times for the models were as follows: T5-small took 24 hours, flant5-small took 30 hours, and BART-base took 50 hours to train for 10 epochs. 

\subsection{Model Inference}
% \subsubsection{Hyperparameters}
% \subsubsection{Post Processing}
Model Inference is a bit more tricky than anticipated since hyperparameters need to be explored to gain the optimal output. The hyperparameters for inference were set differently for each model to optimize their performance.

For BART, the maximum amount of new tokens for a generation was limited to 256, whilst early stopping was enabled to prevent the generation of unnecessary tokens. Sampling was enabled with 100 beams. The top-p value was set to 0.35 to control the nucleus sampling, and the no-repeat parameter for n-gram size was set to 2 so that it could prevent the model from generating repetitive phrases. The temperature was set to 2.5 to control the randomness of the output.

For the T5 model, the restriction on repeating n-grams was configured with a limit of 2. Moreover, the top-k parameter had a configuration value of 50,000 for its sampling method. The setting for temperature was adjusted to 0.7, and for nucleus sampling, the top-p parameter was calibrated to 0.75. Additionally, the model incorporated early stopping.

In the case of flan T5, early stopping was integrated. The configuration for sampling included a limitation on the n-gram repetition, which was established at 2. The beam search was conducted with 200 beams, and the temperature parameter was adjusted to a level of 1.5.

Some post-processing was done to refine the output of the models. Since the output from the models was in lowercase, the first letter of each sentence was capitalized to ensure proper sentence structure. Next, we performed entity case correction. For this, we used the Named Entity Recognition (NER) model from the SpaCy library\footnote{https://github.com/explosion/spaCy}. The NER model identifies various entities within the text, including the names of people, organizations, or locations. We preserved the case capitalization for these entities, ensuring that proper nouns were correctly capitalized in the output. This step is crucial as it enhances the readability and accuracy of the generated sentences. Finally, if multiple sequences were output, we removed any duplicates to ensure the uniqueness of the generated paraphrases. This step helps to maintain the diversity of the output, providing a wider range of paraphrases for each input sentence.

\section{Evaluation}

\begin{table*}
\caption{Comparison of Semantic Similarity of the original dataset, distilled models and teacher model.}
\centering
\renewcommand{\arraystretch}{1.5}
\begin{tabular}{ccccccc}
\hline
\textbf{Model} & \textbf{ADA Score (↑)} & \textbf{SimCSE Score (↑)} & \textbf{PromCSE Score (↑)} & \textbf{Roberta Score (↑)} & \textbf{Mpnet Score (↑)} \\
\hline
Original Dataset & 90.88\% & 71.90\% & 98.68\% & 72.10\% & 69.79\% \\
\hline
ChatGPT & 95.60\% & 91.25\% & 99.41\% & 88.23\% & 87.03\% \\
T5 Small (Ours) & 97.28\% & 94.59\% & 99.67\% & 92.77\% & 92.60\% \\
Flan T5 Small (Ours) & 97.75\% & 95.42\% & 99.71\% & 93.71\% & 93.69\% \\
BART Base (Ours) & \textbf{98.07\%} & \textbf{95.77\%} & \textbf{99.72\%} & \textbf{94.04\%} & \textbf{93.77\%} \\
\hline
\end{tabular}

\label{tab:semantic-comparison}
\end{table*}

\label{sec:evaluation-section}
The evaluation was done using both quantitative and qualitative evaluation techniques. For the quantitative analysis, we utilized 100,000 paraphrase pairs extracted from the MRPC dataset, the MSCOCO evaluation subset, the Twitter URL dataset, and the Wiki Answer dataset. For the qualitative analysis, we incorporated human evaluations and a novel evaluation technique known as LLM evaluations. The subsequent section will outline our findings in the evaluation process.

\subsection{Quantitative Analysis}
For the quantitative analysis, we will be evaluating whether the paraphrases are high quality and diverse. We will assess three main characteristics: semantic similarity, syntactic diversity, and lexical diversity.

\subsubsection{Semantic Simialrity}
In our research, we evaluate semantic similarity by utilizing a range of models to produce embeddings of sentences from both the original text and its paraphrased form. Following this, we measure the degree of similarity using the cosine similarity method between these embeddings. The resulting scores are derived by taking one and subtracting the cosine similarity metric of the base and paraphrased text.

\begin{itemize}
\item The ``Ada Score'' is computed using OpenAI’s text-embedding-ada-002 model \cite{openai_new_2023}.
\item The ``SimCSE Score'' is derived from the sup-simcse-roberta-large model by SimCSE \cite{gao_simcse_2021}.
\item The ``PromCSE Score'' is based on the sup-promcse-roberta-large model from PromCSE \cite{jiang_improved_2022}.
\end{itemize}

We also make use of several models from the sentence-transformers library \cite{reimers_sentence-bert_2019}:

\begin{itemize}
\item The ``Mpnet Score''  is derived by employing the model known as all-mpnet-base-v1.
\item The ``Roberta Score''  is determined through the utilization of the all-roberta-large-v1 model.
\end{itemize}

The data in Table \ref{tab:semantic-comparison} show that despite the distillation process the models have retained the semantic similarity compared to the teacher model. The difference in similarities is negligible likely caused due to different lexical phrases and sentence structures. The original dataset indicates that there is noise in the datasets despite being used in other research for training models.

\subsubsection{Syntactic Diversity}
Syntactic diversity is a measure of the range and intricacy of sentence structures in a paraphrase, given an original sentence. A high level of syntactic diversity suggests that the paraphrased sentences are varied and linguistically sophisticated, meaning that it is a characteristic of a high-quality paraphrase. We evaluate the diversity using metrics that take into account the sentence syntax trees.

\begin{itemize}
\item The ``Ted-F" encompasses the complete Tree Edit Distance measurement. This is achieved through constructing the constituency parse trees for the original and paraphrased sentences using Stanza \cite{qi_stanza_2020}, transforming the trees to bracket notation with the NLTK library \cite{bird_natural_2009} and regex, and then employing the APTED library \cite{pawlik_efficient_2015} to compute the entire Tree Edit Distance.

\item The ``Ted-3" represents Tree Edit Distance of the first three layers. It is calculated in a similar manner to ``Ted-F", but rather than calculating the comprehensive Tree Edit Distance, it focuses on the Tree Edit Distance for the first three strata.

\item The ``Kermit Score" is computed by finding the cosine measurement of similarity between the original and the paraphrase syntactic vectors using the Kermit library \cite{zanzotto_kermit_2020}, and then subtracting this similarity from one. The syntactic embeddings are obtained by feeding the syntax trees of the two sentences.

\item The ``Subtree K Score" is the Subtree Kernel diversity. It is calculated by initially constructing constituency parse trees for the original and paraphrased sentences using Stanza, transforming the trees to an NLTK Tree, and then identifying all the subtrees. The kernel similarity is calculated by the ratio of unique common subtrees to the total count of unique subtrees. This figure is then subtracted from one to yield the diversity score.

\item The ``Node Pair K Score" is the Subtree Node Pair Kernel diversity. It is calculated similarly to the ``Subtree K Score". The only difference is instead of subtrees this uses node pairs for the calculation.
\end{itemize}

Results in Table \ref{tab:syntactic-comparison} show the full results of syntactic diversity. The data suggests that the distilled models have successfully retained the ability to generate syntactically diverse paraphrases similar to the teacher. This characteristic is unseen in most neural-based approaches which is a significant improvement.

\begin{table*}
\caption{Comparison of Syntactic Diversity of the original dataset, distilled models and teacher model.}
\centering
\renewcommand{\arraystretch}{1.5}
\begin{tabular}{cccccc}
\hline
\textbf{Model} & \textbf{Ted-F (↑)} & \textbf{Ted-3 (↑)} & \textbf{Kermit Score (↑)} & \textbf{Subtree K Score (↑)} & \textbf{Node Pair K Score (↑)} \\
\hline
Original Dataset & 17.38 & 4.02 & 74.81\% & 95.36\% & 84.14\% \\
\hline
ChatGPT & 21.24 & 4.53 & \textbf{66.94\%} & \textbf{92.77\%} & \textbf{79.20\%} \\
T5 Small (Ours) & 17.29 & 3.99 & 54.89\% & 83.36\% & 66.58\% \\
Flan T5 Small (Ours) & 18.38 & 4.40 & 54.96\% & 83.23\% & 65.45\% \\
BART Base (Ours) & \textbf{23.45} & \textbf{5.06} & 61.98\% & 88.97\% & 72.30\% \\
\hline
\end{tabular}

\label{tab:syntactic-comparison}
\end{table*}

\begin{table*}
\caption{Comparison of lexical diversity of the original dataset, distilled models and teacher model.}
\centering
\renewcommand{\arraystretch}{1.5}
\begin{tabular}{ccccccc}
\hline
\textbf{Model} & \textbf{BOW Overlap} & \textbf{Corpus BLEU} & \textbf{Corpus BLEU2} & \textbf{METOER} & \textbf{ROUGE 1} & \textbf{ROUGE 2} \\
 & \textbf{Score (↑)} & \textbf{ Score (↑)} & \textbf{ Score(↑)} & \textbf{Score (↑)} & \textbf{Score(↑)} & \textbf{Score(↑)} \\
\hline
Original Dataset & 58.59\% & 99.40\% & 85.12\% & 59.28\% & 53.64\% & 77.02\% \\
\hline
ChatGPT & \textbf{50.55\%} & \textbf{99.54\%} & \textbf{84.01\%} & \textbf{43.20\%} & \textbf{42.56\%} & \textbf{70.15\% }\\
T5 Small (Ours) & 35.42\% & 99.46\% & 65.79\% & 29.50\% & 28.19\% & 50.78\% \\
Flan T5 Small (Ours) & 34.06\% & 99.48\% & 63.93\% & 27.80\% & 26.23\% & 47.05\% \\
BART Base (Ours) & 39.49\% & 99.50\% & 75.71\% & 36.27\% & 33.51\% & 59.79\% \\
\hline
\\
\hline
\textbf{Model} & \textbf{ROUGE L} & \textbf{Token $\cap / \cup$} & \textbf{TER} & \textbf{WER} & \textbf{CharacTER} & \textbf{Google BLEU} \\
 & \textbf{Score (↑)} & \textbf{Score (↑)} & \textbf{Score (↑)} & \textbf{Score (↑)} & \textbf{Score (↑))} & \textbf{Score (↑)} \\
\hline
Original Dataset & 58.49\% & 69.73\% & 80.45 & 85.44 & 69.27 & 80.88\% \\
\hline
ChatGPT & \textbf{54.17\%} & \textbf{62.93\%} & \textbf{63.05} & \textbf{77.44} & \textbf{77.90} & \textbf{77.89\%} \\
T5 Small (Ours) & 42.17\% & 43.52\% & 49.05 & 66.83 & 54.80 & 59.55\% \\
Flan T5 Small (Ours) & 41.98\% & 40.94\% & 48.77 & 69.97 & 56.02 & 57.91\% \\
BART Base (Ours) & 52.11\% & 49.68\% & 59.23 & 82.47 & 68.00 & 68.15\% \\
\hline
\end{tabular}

\label{tab:combined-comparison}
\end{table*}

\subsubsection{Lexical Diversity}
Lexical diversity is a concept that signifies the array and diversity of words incorporated in a text. It serves as an indicator of the width of vocabulary and the application of synonyms. In the sphere of paraphrasing, it becomes essential to gauge lexical diversity to grasp the level of vocabulary fluctuation. We utilized an array of metrics to gauge lexical diversity.

\begin{itemize}
\item ``BOW Overlap Score" is calculated by identifying the shared tokens between the original and the paraphrased text, and dividing by the total count of tokens. This value is then subtracted by one. 

\item ``Corpus BLEU Score" is evaluated using the SacreBLEU Library \cite{post_call_2018}. This score is then subtracted by one.

\item ``Corpus BLEU2 Score" is constructed using the SacreBLEU Library with the “method1” smoothing function. This score is then subtracted by one.

\item ``Sentence BLEU Score" is calculated similarly to the Corpus BLEU score using the SacreBLEU Library but at the sentence level. This score is also subtracted by one.

\item ``METEOR Score" is evaluated using the NLTK library. This score is then subtracted by one.

\item ``ROUGE 1 Score" is constructed using the Google Research library\footnote{{\label{footnote2}https://github.com/google-research/google-research}}. This score is then subtracted by one.

\item ``ROUGE 2 Score" is evaluated using the Google Research library. This score is then subtracted by one.

\item ``ROUGE L Score" is calculated using the Google Research library. This score is then subtracted by one.

\item ``Token $\cap / \cup$ Score" is similar to the BOW Overlap score but with a minor variation. It is calculated using the shared tokens between the original and the paraphrased text, divided by the total unique tokens. This value is then subtracted by one.

\item ``Google BLEU Score" is evaluated using Huggingface’s Evaluate library. This score is then subtracted by one.\footnote{{\label{footnote1}https://github.com/huggingface/evaluate}}

\item ``TER Score" is the Translation Error Rate score which is constructed using Huggingface’s Evaluate library.

\item ``WER Score" is the Word Error Rate score evaluated using Huggingface’s Evaluate library.

\item ``CharacTER Score" is the Character Error Rate score calculated using Huggingface’s Evaluate library.
\end{itemize}

Results in Table \ref{tab:combined-comparison} show the distilled models were able to generate lexically diverse paraphrases on par with the teacher model. Even though the teacher model is better, models still exhibit significant lexical diversity similar to that of the teacher.

\begin{table*}
\caption{Human evaluation Results.}
\centering
 \renewcommand{\arraystretch}{1.4}
\begin{tabular}{ccccc}
\hline
\textbf{Model} & \textbf{Semantic Similarity (↑)} & \textbf{Lexical Diversity (↑)} & \textbf{Syntactic Diversity (↑)} & \textbf{Grammatical Correctness (↑)} \\
\hline
Original Dataset & 3.47 & 3.08 & 2.93 & 4.53 \\
\hline
ChatGPT & \textbf{4.33} & \textbf{3.43} & \textbf{3.26} & \textbf{4.97} \\
T5 Small (Ours) & 4.21 & 3.23 & 2.95 & 4.83 \\
Flan T5 Small (Ours) & 4.05 & 3.19 & 2.83 & 4.74 \\
BART Base (Ours) & 4.20 & 3.25 & 3.16 & 4.89 \\
\hline
\end{tabular}
\label{tab:human-eval-comparison}
\end{table*}

\begin{table*}
\caption{LLM evaluation results.}
\centering
 \renewcommand{\arraystretch}{1.4}
\begin{tabular}{ccccc}
\hline
\textbf{Model} & \textbf{Semantic Similarity (↑)} & \textbf{Lexical Diversity (↑)} & \textbf{Syntactic Diversity (↑)} & \textbf{Grammatical Correctness (↑)} \\
\hline
Original Dataset & 3.40 & 2.83 & 2.88 & 4.37 \\
\hline
ChatGPT & \textbf{4.82} & \textbf{3.01} & \textbf{3.85} & \textbf{4.89} \\
T5 Small (Ours) & 4.51 & 2.48 & 3.20 & 4.68 \\
Flan T5 Small (Ours) & 4.26 & 2.20 & 2.85 & 4.41 \\
BART Base (Ours) & 4.75 & 2.63 & 3.41 & 4.69 \\
\hline
\end{tabular}
\label{tab:llm-eval-comparison}
\end{table*}

\subsection{Qualitative Analysis}

In order to gain an in-depth insight of the performance of our model, we conducted a qualitative analysis using two distinct methods: human evaluation and Large Language Model (LLM) evaluation.

\subsubsection{Human Evaluation}
\label{sec:huma-eval-sec}

We recruited a group of five independent evaluators who are proficient in English for the human evaluation phase. The evaluation process involved the selection of thousand paraphrase pairs from four different data sources: the MRPC dataset, the MSCOCO dataset evaluation subset, the Twitter URL dataset, and the Wiki Answer dataset. From each of these data sources, 250 paraphrase pairs were selected, ensuring a balanced representation. The selection process was carefully conducted to ensure that the lengths of the sentences from each data source were accurately represented. For each of the thousand paraphrase pairs, we obtained model outputs from four different models: ChatGPT, and the three trained models. This resulted in a total annotation set of 5000 pairs for evaluation, which includes both the original dataset paraphrase pairs and the model outputs. The extensive set of evaluations facilitates a comprehensive and reliable assessment of each model's performance.

In our assessment, we employed a 5-point Likert scale \cite{van_der_lee_best_2019} for the evaluation of Semantic Similarity, Lexical Diversity, Syntactic Diversity, and Grammatical Correctness. The breakdown of the Likert scale is as follows:

\begin{itemize}
\item For Semantic Similarity, a score of 5 implies that the text's meaning aligns perfectly or almost perfectly with the source text. This could suggest that the text expresses the same ideas, reaches the same conclusions, or presents the same arguments. Conversely, a score of 1 in semantic similarity would suggest that the text's meaning is entirely different or unrelated to the source text, indicating differing ideas, conclusions, or arguments.

\item For Lexical Diversity, a score of 5 signifies a broad and rich vocabulary. This could suggest that the text employs a diverse array of words, synonyms, and phrases, and avoids repetition. In contrast, a score of 1 in lexical diversity would suggest a limited vocabulary range, indicating that the text uses a small set of words, heavily relies on a few key phrases, or frequently repeats the same words or phrases.

\item For Syntactic Diversity, a score of 5 denotes a high degree of variation in sentence structure. This could suggest that the text employs a variety of sentence types, lengths, and structures, and avoids repetition. Conversely, a score of 1 in syntactic diversity would suggest minimal variation in sentence structure, indicating that the text uses a limited number of sentence types, heavily relies on a few key structures, or frequently repeats the same sentence structures.

\item For Grammatical Correctness, a score of 5 signifies perfect grammar. This could suggest that the text employs correct punctuation, spelling, and syntax, and avoids grammatical errors. Conversely, a score of 1 in grammatical correctness would suggest significant errors that affect comprehension, indicating that the text contains frequent spelling, punctuation, or syntax errors, or that these errors hinder its comprehensibility.
\end{itemize}

The evaluation instruction given to the human evaluators can be seen in Fig.~\ref{fig:human-annotator-instructions}. Grammatical correctness is one aspect that is normally evaluated by other research work but is crucial to identifying the effectiveness of a paraphrase. The final data obtained from the human evaluation are given in Table \ref{tab:human-eval-comparison}

% \begin{table*}
% \caption{Human evaluation Results.}
% \centering
% \begin{tabular}{lcccc}
% \hline
% \textbf{Model} & \textbf{Semantic Similarity (↑)} & \textbf{Lexical Diversity (↑)} & \textbf{Syntactic Diversity (↑)} & \textbf{Grammatical Correctness (↑)} \\
% \hline
% Original Dataset & 3.47 & 3.08 & 2.93 & 4.53 \\
% \hline
% ChatGPT & \textbf{4.33} & \textbf{3.43} & \textbf{3.26} & \textbf{4.97} \\
% T5 Small (Ours) & 4.21 & 3.23 & 2.95 & 4.83 \\
% Flan T5 Small (Ours) & 4.05 & 3.19 & 2.83 & 4.74 \\
% BART Base (Ours) & 4.20 & 3.25 & 3.16 & 4.89 \\
% \hline
% \end{tabular}
% \label{tab:human-eval-comparison}
% \end{table*}

% \begin{table*}
% \caption{LLM evaluation results.}
% \centering
% \begin{tabular}{lcccc}
% \hline
% \textbf{Model} & \textbf{Semantic Similarity (↑)} & \textbf{Lexical Diversity (↑)} & \textbf{Syntactic Diversity (↑)} & \textbf{Grammatical Correctness (↑)} \\
% \hline
% Original Dataset & 3.40 & 2.83 & 2.88 & 4.37 \\
% \hline
% ChatGPT & \textbf{4.82} & \textbf{3.51} & \textbf{3.85} & \textbf{4.89} \\
% T5 Small (Ours) & 4.51 & 2.48 & 3.20 & 4.68 \\
% Flan T5 Small (Ours) & 4.26 & 2.20 & 2.85 & 4.41 \\
% BART Base (Ours) & 4.75 & 2.63 & 3.41 & 4.69 \\
% \hline
% \end{tabular}
% \label{tab:llm-eval-comparison}
% \end{table*}

\subsubsection{LLM Evaluation}

Over the recent year, the application of Language Model Metrics (LLMs) for evaluation in NLP has seen some rise. This surge in popularity is primarily attributed to the superior performance of LLMs over existing reference-free metrics \cite{liu_g-eval_2023}. Recognizing this potential, our research also embarked on an evaluation strategy using LLMs, utilizing OpenAI's gpt-4 model, which was the state-of-the-art (SOTA) LLM at the time of conducting this study \cite{openai_gpt-4_2023}. 

The data used for this evaluation was identical to the one provided to our human annotators. We crafted the prompt in alignment with the instructions given to the human annotators, as depicted in Fig. \ref{fig:llm-eval-prompt}. This approach ensured a fair evaluation strategy to that of the human evaluation.

This innovative methodology employed in our research could serve as a valuable reference for future studies in this field. It not only provides a new perspective on the use of LLMs in NLP but also opens up possibilities for further exploration and development of more advanced and efficient evaluation techniques. The final results of the LLM evaluation are given in Table \ref{tab:llm-eval-comparison}

% \begin{table*}
% \caption{LLM evaluation results.}
% \centering
% \begin{tabular}{lcccc}
% \hline
% \textbf{Model} & \textbf{Semantic Similarity (↑)} & \textbf{Lexical Diversity (↑)} & \textbf{Syntactic Diversity (↑)} & \textbf{Grammatical Correctness (↑)} \\
% \hline
% Original Dataset & 3.40 & 2.83 & 2.88 & 4.37 \\
% \hline
% ChatGPT & \textbf{4.82} & \textbf{3.51} & \textbf{3.85} & \textbf{4.89} \\
% T5 Small (Ours) & 4.51 & 2.48 & 3.20 & 4.68 \\
% Flan T5 Small (Ours) & 4.26 & 2.20 & 2.85 & 4.41 \\
% BART Base (Ours) & 4.75 & 2.63 & 3.41 & 4.69 \\
% \hline
% \end{tabular}
% \label{tab:llm-eval-comparison}
% \end{table*}

\section{Discussion}

Our evaluation process, which combined both quantitative and qualitative methods, provided a comprehensive understanding of the performance of our models. The quantitative analysis focused on three main characteristics: semantic similarity, syntactic diversity, and lexical diversity. The results indicated that, despite the distillation process, our models were able to maintain semantic similarity, syntactic diversity, and lexical diversity in comparison to the teacher model (ChatGPT). 

Our comprehensive qualitative evaluation, which incorporated human evaluation, provided an accurate and in-depth understanding of our research outcomes. The results distinctly demonstrated that, despite our models being a thousand times smaller than ChatGPT, they were able to maintain comparable performance levels. This is a significant achievement, highlighting the efficiency and effectiveness of our models in generating high-quality paraphrases. Moreover, the use of LLM evaluation in our study introduced a novel approach to performance assessment in our research domain. This innovative strategy, which leverages the capabilities of SOTA language models for evaluation, offers a promising avenue for future research. With further refinement and development, this LLM evaluation approach has the potential to serve as a robust benchmarking tool for assessing the performance of paraphrasing models and other natural language processing tasks.

In our evaluation, we employed a wide array of metrics to assess various aspects of the generated paraphrases. This approach was adopted in recognition of the fact that a single metric may not fully capture the effectiveness of a paraphrase, as it tends to focus on a specific characteristic. This highlights an active area of research that warrants further exploration. The use of inappropriate or insufficient metrics can lead to a skewed understanding of the models' performance. Therefore, future research should emphasize the development of a unified metric that can holistically evaluate the quality of paraphrases. Such a metric would not only provide a more precise assessment of the model performance but also contribute to the advancement of the field of paraphrase generation.

In our methodology even though the distilled models were able to retain the quality of the distilled knowledge. One area where the ChatGPT is superior is the ability to generate paraphrases that are diverse from each other. This means that there was a notable variation among the paraphrases generated by ChatGPT. Our models, however, did not guarantee the same level of diversity, indicating a need for further research to optimize the inference hyperparameters to enhance the diversity of the generated paraphrases. Random Sampling could be a potential starting point for this approach, but more extensive work is required to fully address this issue.

Since our models were trained using ChatGPT, they may inherit potential risks associated with it. This could include the propagation of biases inherent in the train dataset. Therefore, it is crucial for future research to address these issues and develop strategies to mitigate the potential risks associated with the use of LLMs with knowledge distillation for paraphrase generation.

\section{Conclusion}
Our research offers a more efficient and cost-effective solution for paraphrase generation, making it more accessible for various applications. The distilled models provide a viable alternative to using large LLMs, opening up new possibilities for their application in real-world scenarios. A notable result of our models is the ability to produce both lexically and syntactically diverse paraphrases. Our models were at least a thousand times smaller than the original LLM and were still able to perform on par with it.

Overall, our work sets the foundation for future advancements in parameter-efficient and diverse paraphrase generation. It  underscores the opportunities for additional exploration and innovation in this domain, paving the way for more efficient and accessible solutions in the field of NLG and paraphrase generation.

\section{Acknowledgment}
We wish to extend our appreciation and thanks to the research participants who participated in the human evaluation of the models. Their insights and feedback have been instrumental in assessing the quality and effectiveness of our research.

\bibliographystyle{IEEEtran}
\bibliography{references}

% \newpage

\onecolumn
% \appendix
\clearpage
\appendix
% \appendices

\renewcommand{\thefigure}{A\arabic{figure}}
\setcounter{figure}{0} % resetting figure counter

\section{Appendix A}
\begin{figure}[H]
\centering
\noindent\fbox{%
    \begin{minipage}{0.85\textwidth}
    \textbf{Source Text:} \$source\_text

    \textbf{Paraphrase:} \$paraphrase

    Please evaluate the following aspects of the paraphrase in comparison to its source text on a likert scale of 1 to 5, where:

    \textbf{Semantic Similarity:} This refers to how closely the meaning of the paraphrase matches the meaning of the source text.

    \textit{Rating Scale for Semantic Similairty}

    1: The paraphrase has a completely different meaning or is unrelated to the source text.

    2: The paraphrase has a somewhat different meaning from the source text

    3: The paraphrase captures the general idea of the source text, but some details or nuances are missing.

    4: The paraphrase largely captures the meaning of the source text but may have slight differences in wording or expression.

    5: The paraphrase has an identical or nearly identical meaning to the source text.

    \textbf{Lexical Diversity:} This aspect evaluates the range and richness of vocabulary used in the paraphrase, considering its comparison to the source text.

    \textit{Rating Scale for Lexical Diversity}

    1: The paraphrase shows a limited use of words and lacks diversity when compared to the source text.

    2: The paraphrase exhibits some variation in word choice but heavily relies on a few specific terms, which may not reflect the lexical diversity of the source text.

    3: The paraphrase demonstrates moderate diversity in vocabulary, but there is room for improvement in terms of incorporating more varied word choices from the source text.

    4: The paraphrase displays a good range of vocabulary, utilizing several different words and expressions that align with the lexical diversity of the source text.

    5: The paraphrase showcases an extensive array of vocabulary, demonstrating excellent lexical diversity that closely matches or surpasses the richness of the source text.

    \textbf{Syntactic Diversity:} This aspect assesses the structural variations in the paraphrase compared to the source text.

    \textit{Rating Scale for Syntactic Diversity}

    1: The paraphrase closely mirrors the sentence structure of the source text with minimal variation.

    2: The paraphrase shows some minor changes in sentence structure but largely follows the same pattern as the source text.

    3: The paraphrase introduces moderate variations in sentence structure, deviating from the structure of the source text in certain aspects.

    4: The paraphrase exhibits significant syntactic diversity, using different sentence structures while still conveying the same meaning as the source text.

    5: The paraphrase displays a high level of syntactic diversity, employing various sentence structures creatively while maintaining the meaning of the source text.

    \textbf{Grammatical Correctness:} This evaluates the grammatical accuracy of the paraphrase.

    \textit{Rating Scale for Grammatical Correctness}

    1: The paraphrase contains numerous grammatical errors that significantly impact comprehension.

    2: The paraphrase has several grammatical errors that occasionally affect understanding.

    3: The paraphrase includes some grammatical errors, but they do not hinder overall comprehension.

    4: The paraphrase demonstrates good grammatical correctness with only occasional minor errors.

    5: The paraphrase is grammatically flawless, with no errors or inaccuracies.

    Please provide your ratings for each aspect using the following json format:

    \{"Semantic Similarity": [Rating from 1 to 5],

    "Lexical Diversity": [Rating from 1 to 5],

    "Syntactic Diversity": [Rating from 1 to 5],

    "Grammatical Correctness": [Rating from 1 to 5]\}
    \end{minipage}%
}
\caption{This figure illustrates the prompt fed to the gpt-4 model for evaluation.}
\label{fig:llm-eval-prompt}
\end{figure}

% \begin{figure}[H]
%     \centering
%     \includegraphics[width=0.88\textwidth]{LLM_prompt.png}
%     \caption{Instructions given to Human Evaluators.}
%     \label{fig:llm-eval-prompt}
% \end{figure}

\section{Appendix B}
\begin{figure}
    \vspace*{-1cm}
    \centering
    \includegraphics[width=0.52\textwidth]{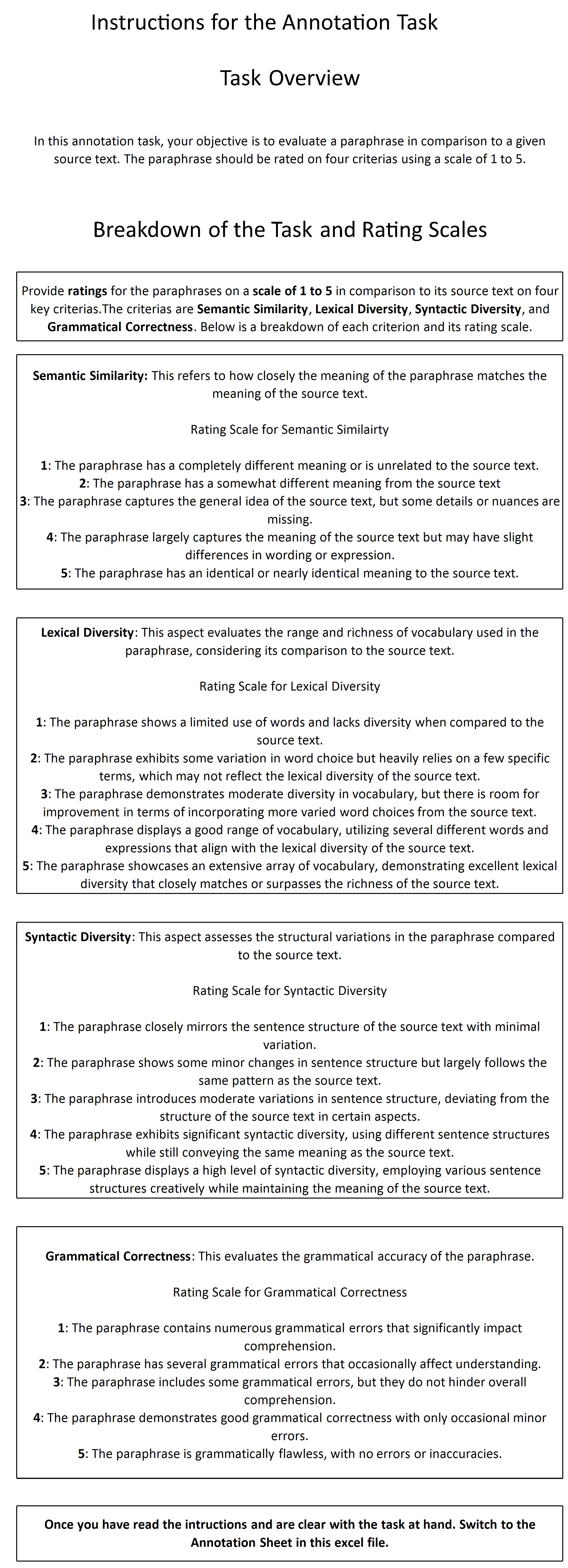}
    \caption{Instructions given to Human Evaluators.}
    \label{fig:human-annotator-instructions}
\end{figure}

\end{document}